\documentclass[10pt, a4paper]{article}

\usepackage{lrec-coling2024} 
\usepackage{times}
\usepackage{latexsym}
\usepackage{tabularx}
\usepackage{nopageno}
\usepackage{booktabs}
\usepackage{graphicx}
\usepackage{multirow}
\usepackage{algorithmic}
\usepackage{tikz}
\usepackage{algorithm2e}
\usetikzlibrary{shapes, arrows.meta, positioning}
\usepackage{tabularx}
\usepackage{booktabs}

\title{Can Similarity-Based Domain-Ordering Reduce \\ Catastrophic Forgetting for Intent Recognition?}

\name{Amogh Mannekote, Xiaoyi Tian, Kristy Elizabeth Boyer, Bonnie J. Dorr}
\address{University of Florida \\
\texttt{\{amogh.mannekote,tianx,keboyer,bonniejdorr\}@ufl.edu}}

\begin{document}
\abstract{
Task-oriented dialogue systems are expected to handle a constantly expanding set of intents and domains even after they have been deployed to support more and more functionalities.
To live up to this expectation, it becomes critical to mitigate the catastrophic forgetting problem (CF) that occurs in continual learning (CL) settings for a task such as intent recognition.
While existing dialogue systems research has explored replay-based and regularization-based methods to this end, the effect of domain ordering on the CL performance of intent recognition models remains unexplored.
If understood well, domain ordering has the potential to be an orthogonal technique that can be leveraged alongside existing techniques such as experience replay.
Our work fills this gap by comparing the impact of three domain-ordering strategies (min-sum path, max-sum path, random) on the CL performance of a generative intent recognition model.
Our findings reveal that the min-sum path strategy outperforms the others in reducing catastrophic forgetting when training on the 220M T5-Base model.
However, this advantage diminishes with the larger 770M T5-Large model.
These results underscores the potential of domain ordering as a complementary strategy for mitigating catastrophic forgetting in continually learning intent recognition models, particularly in resource-constrained scenarios.
\newline \Keywords{continual learning, slot filling, intent classification}
}
\maketitleabstract

\thispagestyle{empty}

\section{Introduction}

In real-world dialogue systems, language use can vary significantly across different domains or contexts. For example, a dialogue model used in a customer service dialogue system often needs to be dynamically repurposed to handle new types of inquiries or new product offerings after deployment. However, it can be expensive to retrain the underlying dialogue model from scratch every time there is such a change. For this reason, continual learning (CL, or lifelong learning) has become a popular paradigm for training task-oriented dialogue systems \cite{ke2022continuallearning,lee2017toward,li2022continual}.

CL-based approaches to update existing dialogue models enable them to adapt to constantly evolving business needs without the need for regular, full-retraining from scratch, making them more resource-efficient in terms of compute, energy, and time \cite{van2019three,chen2018lifelong}. This paper focuses on \textit{intent recognition}, which a crucial early step in pipeline-based dialogue systems, affecting downstream modules like dialogue state tracking, dialogue management, and natural language generation.

\begin{figure}[t]
    \centering
    \includegraphics[scale = 0.35]{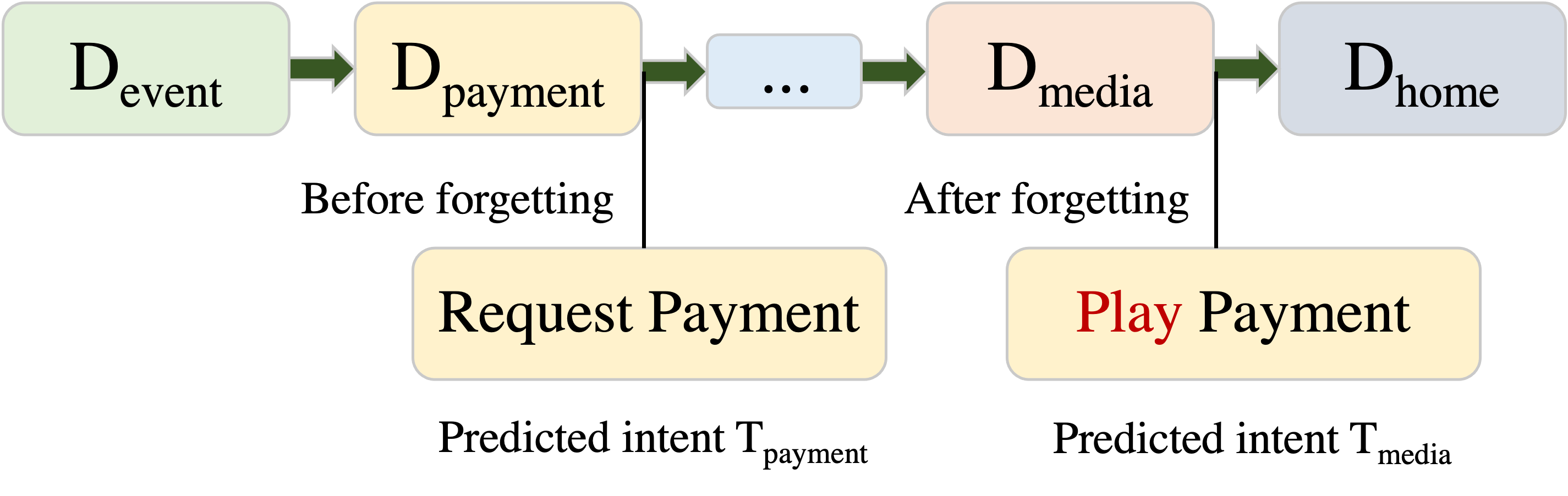}
    \caption{In continual learning, the model is trained one domain at the time. In this example, the model is first trained on data from the event domain $D_{event}$, then trained based on the payment domain $D_{payment}$ and so on.  ``request payment'' might be mislabeled as ``play payment'' after $D_{media}$ training due to CF.}
    \label{fig:forgetting-example}
\end{figure}

Catastrophic forgetting (CF) is a significant challenge in CL settings, where the model's performance deteriorates on previously learned tasks as it encounters new ones \cite{nguyen2019towardunderstanding,cf1970article,mccloskey1989catastrophic}. This is especially prominent in multi-domain intent recognition. When the model is trained incrementally on one domain at a time (as depicted in Figure \ref{fig:forgetting-example}), it is susceptible to forgetting when trained on subsequent domains. Prior research has explored mitigating strategies such as regularization \cite{li2022overcomingcatastrophic}, experience replay \cite{qin2022elle}, and curriculum learning \cite{bengio2009curriculum}.


Domain ordering, rooted in the principles of curriculum learning, has been effective in mitigating CF in continual learning \cite{shui2019aprincipledapproach}. Curriculum learning suggests that learning is more effective when examples progress from simple to complex, mirroring how humans and animals learn. The curriculum approach sorts the dialogue samples by difficulty, then guides the model to learn in increasing order of the ``complexity'' of the samples (with varying definitions of complexity, of course) \cite{dai2021preview, wang2021survey}. While task-ordering effects on binary image classifications \cite{shui2019aprincipledapproach} using MNIST dataset \cite{lecun1998mnist} have been explored, there is limited investigation on its impact on dialogue-based tasks. 


In the context of task-oriented dialogue systems, \citet{madotto2021continuallearning} explored techniques to improve continual learning performance for various tasks such as natural language understanding, dialogue state tracking, and natural language generation. Similarly, \citet{lee2017toward} and \citet{mi2020continual} deployed regularization techniques such as Elastic Weight Consolidation (EWC) for natural language generation as well as end-to-end dialogue setups. We follow in the footsteps of these prior works and investigate how domain ordering influences CF in continual learning for intent recognition.





Our study addresses two research questions regarding intent recognition in a dialogue setting:
\begin{enumerate}
    \item Does the extent of CF in a continually trained intent recognition model depend on the order in which the domains are presented?
    \item If the sequence does impact CF, which ordering strategy is most effective in reducing it?
\end{enumerate}


To explore these questions, we employ a three-step process for our experiments using the Schema-Guided Dialog (SGD) corpus \citeplanguageresource{48569}. First, we extract fixed-size subsets from the power-set of all the domains in the corpus. We then arrange each subset using three ordering strategies (minimum sum path, maximum sum path, and random domain ordering) that based on inter-domain similarities. In the final step, we continually train models over these subsets using different model sizes (T5-Base and T5-Large) and evaluate their performance using \textit{overall accuracy} and \textit{catastrophic forgetting} as evaluation metrics.

Note that our study does not aim to compare performance differences across model sizes; it is widely accepted that a larger model with more parameters will inherently yield better results \cite{raffel2020exploring}. Instead, our research primarily investigates whether different domain ordering strategies influence performance outcomes under the same computational constraints.

\section{Methods} \label{sec:methods}


At a high-level, we generate domain orderings using a three-step procedure.
\begin{enumerate}
    \item We first calculate an inter-domain distance matrix by calculating the mean cosine similarity between the S-BERT embeddings \cite{reimers-gurevych-2019-sentence} of the utterances in each domain.
    \item Next, we construct a domain-similarity graph, which is a fully-connected, weighted, undirected graph whose each node represents a domain and the weight of the edge connecting two nodes is equal to the inter-domain distance between their corresponding domains.
    \item Finally, given a domain-similarity graph, we calculate three separate domain orderings of those domains. We formulate these orderings as various path-finding algorithms (in our case, we examine three paths: min-path, max-path, and random paths).
\end{enumerate}

\subsection{Computing Domain Orderings}

We first randomly sample five domains from all available domains in our dialogue corpus.
Then, for each sample, we generate three distinct domain orderings over the domain similarity graph using a brute-force approach. The three domain orderings include: 1) maximum-sum hamiltonian path (max-sum-path), 2) minimum sum hamiltonian path (min-sum-path), and a 3) random path\footnote{Although the general problem of computing the min-sum path and max-sum path is NP-hard \cite{10.5555/1614191}, we keep the problem size (the number of domains in a single run) small enough such that a brute-force approach is computationally feasible.}.
These ordering strategies have been utilized in previous work on task-ordering in continual learning such as \citet{bell2022theeffectof}. However, it is significantly different from our setting as their experiments are run on the MNIST image dataset \cite{LeCun2005TheMD} for a binary classification task.

Finally, we train the model based on the ordering strategies listed above.
In total, we sample 22 domain subsets (five domains) and obtain 66 continual learning runs (three unique orderings per domain subset).
\begin{figure}[h]
    \centering
    \includegraphics[scale=0.65]{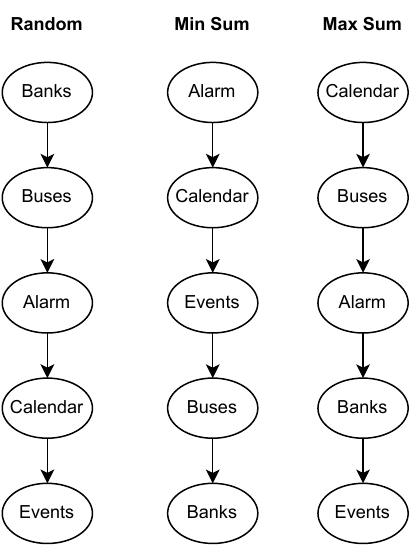}
    \caption{Random, Min-Sum Path, and Max-Sum Path ordering strategies.}
    \label{fig:min_sum_max_sum}
\end{figure}





\subsection{Evaluation Metrics}
We evaluate a model's continual learning performance using two metrics: \textbf{Average Accuracy} and \textbf{Average CF}. These metrics have been used widely in previous studies related to continual learning \cite{bell2022theeffectof,wang2022learning,chaudhry2018efficient}. Average Accuracy is defined as the mean accuracy achieved by the final model over all domains, while Average CF is the average difference between the best accuracy achieved during training and the final accuracy for all tasks (except the last task).




\section{Experiments} \label{sec:experiments}

\paragraph{Dataset.} Our experiments are conducted using the Schema-Guided Dialog (SGD) dataset \cite{48569}, which contains over 20,000 dialogues and covers 20 domains (each domain has anywhere from 1 to 10 intents).
The test split of the dataset contains unseen domains to facilitate evaluation of out-of-domain generalization performance. Figure~\ref{fig:min_sum_max_sum} shows the three domain orderings for the top five domains in the SGD dataset.

\paragraph{Tokenization.}
We prefix the model input with \texttt{``classify intent: ''}. To this we concatenate the dialogue history by combining all utterances within a fixed dialogue context window. Individual utterances in the dialogue history are separated by the separator token used for T5 models, \texttt{</s>}. To prompt the model to predict the intent label, we add \texttt{``intent: ''} at the end of the input sequence. The input is truncated to a maximum of 512 tokens, and we set the context window size to 3. For the output label, we retain the original labels (e.g., \texttt{MakeBooking}). 

\paragraph{Training.}
We utilize T5-based models (T5-base and T5-large) as our backbone models for all our experiments \cite{raffel2020exploring}.
We incrementally fine-tune the models over each domain one by one.
We limit the total number of samples per domain to 100,
with 60\% allocated to training, 
15\% allocated for validation,
and 25\% for testing. For each T5 model (\textsc{T5-Base} and \textsc{T5-Large}), we perform 22 runs for each ordering strategy (total of 66 runs).

\section{Results} \label{sec:results}

\begin{figure*}[ht]
\centering
 \includegraphics[width=\textwidth]{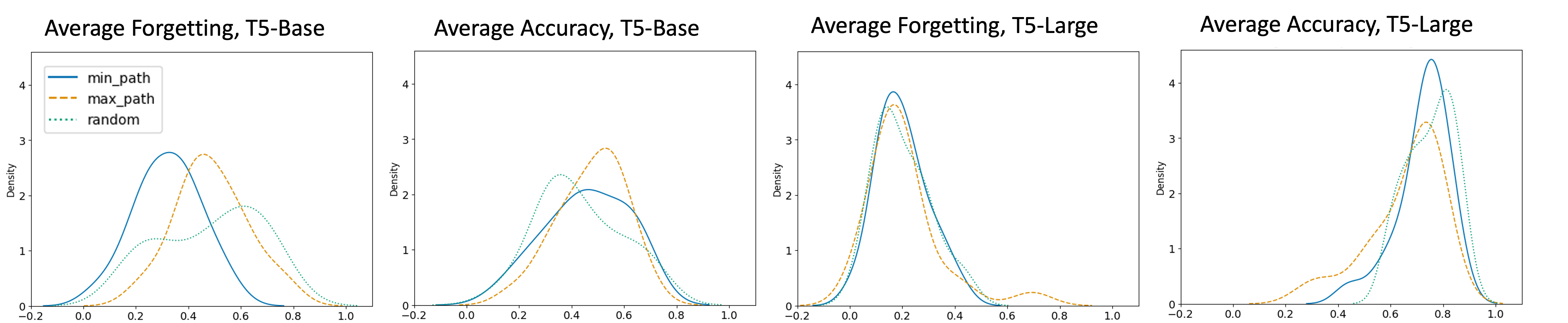}
    \caption{Density plots of Average Accuracy and Average CF (Forgetting) for min-sum path, max-sum path and random ordering strategies using T5-Base and T5-Large models across 22 domain subsets. Min-sum-path strategy with \textsc{T5-Base} results in minimal Average CF, while no significant differences are observed in Average Accuracy between strategies or with \textsc{T5-Large} models.}
   \label{fig:density-plots}
\end{figure*}

Figure \ref{fig:density-plots} 
compares the distributions of Average Accuracy and Average CF among the three Ordering Strategies based on 22 runs.
The outcomes of these different ordering strategies are summarized in Table \ref{tab:results}.
To evaluate the impact of domain ordering strategies on model performance, we perform a repeated-measures ANOVA. We intentionally do not draw comparisons between the performance of different T5 model sizes.
Instead, we explore the impact of various domain ordering strategies on model performance, all while keeping computing resources constant. 


\begin{table*}[h]
\small
\centering
\caption{Results for each ordering strategy for T5-Base and T5-Large models}
\label{tab:evaluation_metrics}
\begin{tabular}{lcccccccc}
\toprule
Model & \multicolumn{4}{c}{T5-Base} & \multicolumn{4}{c}{T5-Large} \\
\cmidrule(lr){2-5} \cmidrule(lr){6-9}
Metric & \multicolumn{2}{c}{Average Accuracy} & \multicolumn{2}{c}{Average CF} & \multicolumn{2}{c}{Average Accuracy} & \multicolumn{2}{c}{Average CF} \\
\cmidrule(lr){2-3} \cmidrule(lr){4-5} \cmidrule(lr){6-7} \cmidrule(lr){8-9}
Ordering Strategy & Mean & SD & Mean & SD & Mean & SD & Mean & SD \\
\midrule
min-sum path & 0.457 & 0.156 & 0.321 & 0.125 & 0.724 & 0.1 & 0.199 & 0.097 \\
max-sum path & 0.476 & 0.128 & 0.489 & 0.136 & 0.657 & 0.141 & 0.205 & 0.141 \\
random & 0.431 & 0.16 & 0.484 & 0.192 & 0.751 & 0.088 & 0.201 & 0.105 \\
\midrule
ANOVA results & \multicolumn{2}{c}{\textit{F} = 0.5004, \textit{p} = 0.6086} & \multicolumn{2}{c}{\textbf{\textit{F} = 8.476, \textit{p} = 0.0006}} & \multicolumn{2}{c}{\textit{F} = 3.293, \textit{p} = 0.077} & \multicolumn{2}{c}{\textit{F} = 0.028, \textit{p} = 0.869} \\
\bottomrule
\end{tabular}
\label{tab:results}
\end{table*}

The ANOVA results show that, for the \textbf{T5-Base model}, there is a statistically significant difference in Average CF between the ordering strategies (F(2, 63) = 8.476, p = 0.0006). To delve deeper into these differences, we employ Tukey's HSD Test for multiple comparisons among the three ordering strategies: max-sum path, min-sum path, and random. The test reveals that min-sum path significantly outperforms the max-sum path strategy (p = 0.0017, 95\% C.I. = [0.0565, 0.2793]) as well as the random strategy (p = 0.0024, 95\% C.I. = [0.0515, 0.2743]). Yet, there is no significant difference in Average CF between the max-sum path and random strategies (p = 0.99). Mean and standard deviation for each strategy can be found in Table \ref{tab:results}. Furthermore, the repeated measures ANOVA indicates no significant influence of the ordering strategy on Average Accuracy across the three conditions [F(2, 63) = 0.5004, p = 0.6086]. 

As for the \textbf{T5-Large model}, the ANOVA results indicate that the ordering strategy did not have a significant effect on either Average Accuracy (F(2, 63) = 3.293, p = 0.077) or Average CF (F(2, 63) = 0.028, p = 0.869).
\paragraph{Qualitative Analysis.} Given that our model uses a generative text-to-text approach to generate its intent labels, we observe that both our models are particularly sensitive to how these labels are formed. Specifically, the intent labels from our SGD corpus adhere to a \texttt{<verb>\_<noun>} pattern (e.g., \textit{play\_movie, set\_alarm, book\_hotel}). We find that both T5-Base and T5-Large correctly predict the nouns most of the time. However, the verbs are frequently mispredicted.
For example, the label \textit{play\_movie} is often incorrectly predicted as \textit{reserve\_movie}.
Similarly, \textit{book\_hotel} is incorrectly predicted as \textit{get\_hotel}.
We attribute the relatively low difficulty of identifying the noun to the presence of easily available cues such as keywords present in the utterance. In contrast, the verb is a more ``latent'' factor that needs higher model capacity to recognize. In other words, we find that the catastrophic forgetting is primarily due to the low model capacity in differentiating the verbs.


\section{Discussion} \label{sec:discussion}
Our most important finding is that while the ordering of the training domains can have a significant effect on catastrophic forgetting (CF) during continual learning for smaller model sizes, this effect significantly breaks down when a larger model is used.

Within the smaller model regime (in our case, the smaller model is T5-Base), our results suggest that the \textit{min-sum path} strategy is significantly more effective than the \textit{max-sum path} and \textit{random} ordering strategies.
This finding aligns with curriculum learning concepts \cite{dai2021preview,wang2021survey,xu2020curriculumlearning}, suggesting benefits from progressively increasing the ``jumps'' in dissimilarity between successive training samples.
One explanation for this phenomenon is that this strategy capitalizes on inherent domain similarities, helping the model recognize shared patterns and features. This fortifies its grasp on common concepts, aiding generalization and minimizing CF. Consequently, this would aid the model in becoming more proficient at managing later-introduced, related domains.

In terms of practical implications, our findings suggest that the ordering strategy might be particularly helpful in scenarios with limited computing resources where training larger language models may not always be feasible. In such cases, leveraging an appropriate ordering strategy, such as our identified \textit{min-sum path}, can prove to be a practical method to achieve effective intent recognition while optimizing resource utilization.

\section{Conclusion} \label{sec:conclusion}
Our study investigates the impact of domain ordering on mitigating catastrophic forgetting in intent recognition models. We find that the \textit{min-sum path} strategy, which follows a progression of similar domains, is the most effective in reducing catastrophic forgetting in the smaller, T5-Base model. However, this effect is not observed in the larger T5-Large model, suggesting that model size and complexity may influence the importance of domain ordering. These findings highlight the importance of carefully considering domain ordering strategies, particularly in resource-constrained scenarios.

\section*{Limitations}

The first limitation of this approach is that it is a strategy based on domain ordering. Therefore, it is inherently only applicable to those settings where the model training does not need to occur one-by-one. 
The second limitation of our work involves the use of a singular definition of inter-domain distance. There are several possible ways to define inter-domain distance, including those based on notions of ``difficulty'' of a domain.
Finally, as is often true with preliminary experiments such as this one, future studies should incorporate larger and more diverse datasets to enhance generalizability of the results.


\section*{Ethics Statement on Broader Impact}
By improving the performance and stability of dialogue-based systems, this work holds the potential to give rise to more efficient and accurate AI technologies. Improvements in the continual learning model performance of intent recognition models, lead to reduction of compute and time required to update widely-deployed NLU models in dialogue systems. With rising global adoption of virtual assistants such as Amazon Alexa and Google Assistant, improvements in efficient training can help reduce carbon emissions from commercial data centers, particularly in the face of ever-increasing sizes of large language models continue to increase. In terms of data ethics, our study uses the publicly-available Schema-Guided Dialog (SGD) corpus, which is in standard use for evaluating dialogue models. We did not collect any new datasets for this exploration.

\bibliographystyle{lrec-coling2024-natbib}
\bibliography{custom}

\begin{thebibliography}{25}
\expandafter\ifx\csname natexlab\endcsname\relax\def\natexlab#1{#1}\fi

\bibitem[{Bell and Lawrence(2022)}]{bell2022theeffectof}
Samuel~J. Bell and Neil~D. Lawrence. 2022.
\newblock \href {https://arxiv.org/abs/2205.13323} {The effect of task ordering
  in continual learning}.
\newblock \emph{arXiv preprint arXiv:2205.13323}.

\bibitem[{Bengio et~al.(2009)Bengio, Louradour, Collobert, and
  Weston}]{bengio2009curriculum}
Yoshua Bengio, J{\'e}r{\^o}me Louradour, Ronan Collobert, and Jason Weston.
  2009.
\newblock Curriculum learning.
\newblock In \emph{Proceedings of the 26th annual international conference on
  machine learning}, pages 41--48.

\bibitem[{Chaudhry et~al.(2018)Chaudhry, Ranzato, Rohrbach, and
  Elhoseiny}]{chaudhry2018efficient}
Arslan Chaudhry, Marc'Aurelio Ranzato, Marcus Rohrbach, and Mohamed Elhoseiny.
  2018.
\newblock Efficient lifelong learning with a-gem.
\newblock \emph{arXiv preprint arXiv:1812.00420}.

\bibitem[{Chen and Liu(2018)}]{chen2018lifelong}
Zhiyuan Chen and Bing Liu. 2018.
\newblock Lifelong machine learning.
\newblock \emph{Synthesis Lectures on Artificial Intelligence and Machine
  Learning}, 12(3):1--207.

\bibitem[{Cormen et~al.(2009)Cormen, Leiserson, Rivest, and
  Stein}]{10.5555/1614191}
Thomas~H. Cormen, Charles~E. Leiserson, Ronald~L. Rivest, and Clifford Stein.
  2009.
\newblock \emph{Introduction to Algorithms, Third Edition}, 3rd edition.
\newblock The MIT Press.

\bibitem[{Dai et~al.(2021)Dai, Li, Li, Sun, Huang, Si, and
  Zhu}]{dai2021preview}
Yinpei Dai, Hangyu Li, Yongbin Li, Jian Sun, Fei Huang, Luo Si, and Xiaodan
  Zhu. 2021.
\newblock Preview, attend and review: Schema-aware curriculum learning for
  multi-domain dialog state tracking.
\newblock \emph{arXiv preprint arXiv:2106.00291}.

\bibitem[{French(1970)}]{cf1970article}
Robert French. 1970.
\newblock Using semi-distributed representations to overcome catastrophic
  forgetting in connectionist networks.

\bibitem[{Ke and Liu(2022)}]{ke2022continuallearning}
Zixuan Ke and Bing Liu. 2022.
\newblock \href {https://arxiv.org/abs/2211.12701} {Continual learning of
  natural language processing tasks: A survey}.
\newblock \emph{arXiv preprint arXiv:2211.12701}.

\bibitem[{LeCun(1998)}]{lecun1998mnist}
Yann LeCun. 1998.
\newblock The {MNIST} database of handwritten digits.
\newblock \emph{http://yann.lecun.com/exdb/mnist/}.

\bibitem[{LeCun and Cortes(2005)}]{LeCun2005TheMD}
Yann LeCun and Corinna Cortes. 2005.
\newblock \href {https://api.semanticscholar.org/CorpusID:60282629} {The mnist
  database of handwritten digits}.

\bibitem[{Lee(2017)}]{lee2017toward}
Sungjin Lee. 2017.
\newblock Toward continual learning for conversational agents.
\newblock \emph{arXiv preprint arXiv:1712.09943}.

\bibitem[{Li et~al.(2022{\natexlab{a}})Li, Chen, Cho, Hao, Liu, Xing, Guo, and
  Liu}]{li2022overcomingcatastrophic}
Dingcheng Li, Zheng Chen, Eunah Cho, Jie Hao, Xiaohu Liu, Fan Xing, Chenlei
  Guo, and Yang Liu. 2022{\natexlab{a}}.
\newblock Overcoming catastrophic forgetting during domain adaptation of
  seq2seq language generation.
\newblock In \emph{Proceedings of the 2022 Conference of the North American
  Chapter of the Association for Computational Linguistics: Human Language
  Technologies}, pages 5441--5454.

\bibitem[{Li et~al.(2022{\natexlab{b}})Li, Zhai, Chen, Gao, Zhang, and
  Zhang}]{li2022continual}
Guodun Li, Yuchen Zhai, Qianglong Chen, Xing Gao, Ji~Zhang, and Yin Zhang.
  2022{\natexlab{b}}.
\newblock Continual few-shot intent detection.
\newblock In \emph{Proceedings of the 29th International Conference on
  Computational Linguistics}, pages 333--343.

\bibitem[{Madotto et~al.(2021)Madotto, Lin, Zhou, Moon, Crook, Liu, Yu, Cho,
  Fung, and Wang}]{madotto2021continuallearning}
Andrea Madotto, Zhaojiang Lin, Zhenpeng Zhou, Seungwhan Moon, Paul Crook, Bing
  Liu, Zhou Yu, Eunjoon Cho, Pascale Fung, and Zhiguang Wang. 2021.
\newblock \href {https://aclanthology.org/2021.emnlp-main.590/} {Continual
  learning in task-oriented dialogue systems}.
\newblock \emph{arXiv preprint arXiv:2012.15504}.

\bibitem[{McCloskey and Cohen(1989)}]{mccloskey1989catastrophic}
Michael McCloskey and Neal~J Cohen. 1989.
\newblock Catastrophic interference in connectionist networks: The sequential
  learning problem.
\newblock In \emph{Psychology of learning and motivation}, volume~24, pages
  109--165. Elsevier.

\bibitem[{Mi et~al.(2020)Mi, Chen, Zhao, Huang, and Faltings}]{mi2020continual}
Fei Mi, Liangwei Chen, Mengjie Zhao, Minlie Huang, and Boi Faltings. 2020.
\newblock Continual learning for natural language generation in task-oriented
  dialog systems.
\newblock \emph{arXiv preprint arXiv:2010.00910}.

\bibitem[{Nguyen et~al.(2019)Nguyen, Achille, Lam, Hassner, Mahadevan, and
  Soatto}]{nguyen2019towardunderstanding}
Cuong~V Nguyen, Alessandro Achille, Michael Lam, Tal Hassner, Vijay Mahadevan,
  and Stefano Soatto. 2019.
\newblock Toward understanding catastrophic forgetting in continual learning.
\newblock \emph{arXiv preprint arXiv:1908.01091}.

\bibitem[{Qin et~al.(2022)Qin, Zhang, Lin, Liu, Li, Sun, and
  Zhou}]{qin2022elle}
Yujia Qin, Jiajie Zhang, Yankai Lin, Zhiyuan Liu, Peng Li, Maosong Sun, and Jie
  Zhou. 2022.
\newblock {ELLE}: Efficient lifelong pre-training for emerging data.
\newblock \emph{arXiv preprint arXiv:2203.06311}.

\bibitem[{Raffel et~al.(2020)Raffel, Shazeer, Roberts, Lee, Narang, Matena,
  Zhou, Li, and Liu}]{raffel2020exploring}
Colin Raffel, Noam Shazeer, Adam Roberts, Katherine Lee, Sharan Narang, Michael
  Matena, Yanqi Zhou, Wei Li, and Peter~J Liu. 2020.
\newblock Exploring the limits of transfer learning with a unified text-to-text
  transformer.
\newblock \emph{The Journal of Machine Learning Research}, 21(1):5485--5551.

\bibitem[{Reimers and Gurevych(2019)}]{reimers-gurevych-2019-sentence}
Nils Reimers and Iryna Gurevych. 2019.
\newblock \href {https://doi.org/10.18653/v1/D19-1410} {Sentence-{BERT}:
  Sentence embeddings using {S}iamese {BERT}-networks}.
\newblock In \emph{Proceedings of the 2019 Conference on Empirical Methods in
  Natural Language Processing and the 9th International Joint Conference on
  Natural Language Processing (EMNLP-IJCNLP)}, pages 3982--3992, Hong Kong,
  China. Association for Computational Linguistics.

\bibitem[{Shui et~al.(2019)Shui, Abbasi, Émile Robitaille, Wang, and
  Gagné}]{shui2019aprincipledapproach}
Changjian Shui, Mahdieh Abbasi, Louis Émile Robitaille, Boyu Wang, and
  Christian Gagné. 2019.
\newblock \href {https://arxiv.org/abs/1903.09109} {A principled approach for
  learning task similarity in multitask learning}.
\newblock \emph{arXiv preprint arXiv:1903.09109}.

\bibitem[{Van~de Ven and Tolias(2019)}]{van2019three}
Gido~M Van~de Ven and Andreas~S Tolias. 2019.
\newblock Three scenarios for continual learning.
\newblock \emph{arXiv preprint arXiv:1904.07734}.

\bibitem[{Wang et~al.(2021)Wang, Chen, and Zhu}]{wang2021survey}
Xin Wang, Yudong Chen, and Wenwu Zhu. 2021.
\newblock A survey on curriculum learning.
\newblock \emph{IEEE Transactions on Pattern Analysis and Machine
  Intelligence}, 44(9):4555--4576.

\bibitem[{Wang et~al.(2022)Wang, Zhang, Lee, Zhang, Sun, Ren, Su, Perot, Dy,
  and Pfister}]{wang2022learning}
Zifeng Wang, Zizhao Zhang, Chen-Yu Lee, Han Zhang, Ruoxi Sun, Xiaoqi Ren,
  Guolong Su, Vincent Perot, Jennifer Dy, and Tomas Pfister. 2022.
\newblock Learning to prompt for continual learning.
\newblock In \emph{Proceedings of the IEEE/CVF Conference on Computer Vision
  and Pattern Recognition}, pages 139--149.

\bibitem[{Xu et~al.(2020)Xu, Zhang, Mao, Wang, Xie, and
  Zhang}]{xu2020curriculumlearning}
Benfeng Xu, Licheng Zhang, Zhendong Mao, Quan Wang, Hongtao Xie, and Yongdong
  Zhang. 2020.
\newblock \href {https://doi.org/10.18653/v1/2020.acl-main.542} {Curriculum
  learning for natural language understanding}.
\newblock In \emph{Proceedings of the 58th Annual Meeting of the Association
  for Computational Linguistics}, pages 6095--6104, Online. Association for
  Computational Linguistics.

\end{thebibliography}


\begin{thebibliography}{1}
\expandafter\ifx\csname natexlab\endcsname\relax\def\natexlab#1{#1}\fi

\bibitem[{Rastogi et~al.(2020)Rastogi, Zang, Sunkara, Gupta, and
  Khaitan}]{48569}
Rastogi, Abhinav and Zang, Xiaoxue and Sunkara, Srinivas and Gupta, Raghav and
  Khaitan, Pranav. 2020.
\newblock \emph{Towards scalable multi-domain conversational agents: The
  schema-guided dialogue dataset}.
\newblock Proceedings of the AAAI conference on artificial intelligence.

\end{thebibliography}

\section*{Language Resource References}
\label{lr:ref}
\bibliographystylelanguageresource{lrec-coling2024-natbib}
\bibliographylanguageresource{languagerec}

\end{document}